%% file: document.tex
\documentclass[lettersize,journal]{IEEEtran}[10pt]
\usepackage{amsmath,amsfonts}
\usepackage{algorithmic}
\usepackage{algorithm}
\usepackage{array}
\usepackage[caption=false,font=normalsize,labelfont=sf,textfont=sf]{subfig}
\usepackage{textcomp}
\usepackage{stfloats}
\usepackage{url}
\usepackage{verbatim}
\usepackage{graphicx}
\usepackage{xcolor}
\usepackage{cite}
\usepackage{multirow}
\usepackage{booktabs}
\usepackage{array}
\usepackage[nolist]{acronym}

\usepackage{pifont}% http://ctan.org/pkg/pifont
\newcommand{\cmark}{\ding{51}}%
\newcommand{\xmark}{\ding{55}}%

\begin{document}

\title{
DivQAT: Enhancing Robustness of Quantized 
Convolutional Neural
Networks
% CNNs 
against Model Extraction Attacks
% % Enhancing Robustness of Edge Deep Learning Models: 
% DivQAT: 
% % Divergence-Based Quantization Aware Training as a Defense against Model Extraction Attacks
% % on Quantized CNNs
% Protecting Quantized CNNs
% against Model Extraction Attacks
% through Divergence-Based Quantization Aware Training
% % Div-Q: Diverging Quantization as a Defense Against Extraction Attacks on Quantized DNNs
}

\author{
% Anonymous authors
Kacem Khaled, 
Felipe Gohring de Magalhães and
Gabriela Nicolescu\\
\textit{Polytechnique Montreal}, Canada %~\IEEEmembership{Staff,~IEEE,} 
        % <-this % stops a space
\thanks{
This research was funded by Synopsys Inc. and the Natural Sciences and Engineering Research Council of Canada (NSERC).
}% <-this % stops a space
% \thanks{Manuscript received April 19, 2021; revised August 16, 2021.}
% }
\thanks{Preprint under review.}
}

% The paper headers
% \markboth{Journal of \LaTeX\ Class Files,~Vol.~14, No.~8, August~2021}%
% {Shell \MakeLowercase{\textit{et al.}}: A Sample Article Using IEEEtran.cls for IEEE Journals}

% \IEEEpubid{0000--0000/00\$00.00~\copyright~2021 IEEE}
% Remember, if you use this you must call \IEEEpubidadjcol in the second
% column for its text to clear the IEEEpubid mark.

\maketitle

\begin{abstract}
% Quantization is widely used to reduce the computational costs and memory footprint of \ac{CNNs} in edge devices.
% Model owners often provide the user with API access to query their model.
% Even in a black-box setting, a malicious user can exploit the predictions provided by the model to perform an extraction attack, resulting in a surrogate model that imitates the victim's functionality. 
\ac{CNNs} and their quantized counterparts are vulnerable to extraction attacks, posing a significant threat of IP theft. Yet, the robustness of quantized models against these attacks is little studied compared to large models. 
Previous defenses propose to inject calculated noise into the prediction probabilities. 
However, these defenses are limited since they are not incorporated during the model design and are only added as an afterthought after training. 
Additionally, most defense techniques are computationally expensive and often have unrealistic assumptions about the victim model that are not feasible in edge device implementations and do not apply to quantized models. 
In this paper, we propose \emph{DivQAT}, a novel algorithm to train quantized \ac{CNNs} based on \ac{QAT} aiming to enhance their robustness against extraction attacks.  
To the best of our knowledge, our technique is the first to modify the quantization process to integrate a model extraction defense into the training process. 
Through empirical validation on benchmark vision datasets,
we demonstrate the efficacy of our technique in defending against model extraction attacks without compromising model accuracy. 
Furthermore, combining our quantization technique with other defense mechanisms improves their effectiveness compared to traditional \ac{QAT}.
\end{abstract}


\newcommand\kacem[1]{\textcolor{blue}{#1}}
\newcommand\mouna[1]{\textcolor{cyan}{#1}}

\newcommand\rewrite[1]{\textcolor{gray}{#1}}

% Acronyms
\input{src/acro}
% Main file with technical content
\input{src/0-main}

\bibliography{IEEEabrv,document}
\bibliographystyle{IEEEtran}			% Bibliography style. 

\vspace{11pt}

\end{document}

%% file: src/acro.tex
% List of acronyms and abbreviations
\begin{acronym}
  \acro{ML}{Machine Learning}
  \acro{DL}{Deep Learning}
  \acro{MLaaS}{Machine Learning as a Service}
  \acro{ANN}{Artificial Neural Networks}
  \acro{API}{Application Programming Interface}
  \acro{CNN}{Convolutional Neural Network}
  \acro{CNNs}{Convolutional Neural Networks}
  \acro{DNN}{Deep Neural Network}
  \acro{DNNs}{Deep Neural Networks}
  \acro{NN}{Neural Network}
  \acro{NNs}{Neural Networks}
  \acro{ART}{Adversarial Robustness Toolbox }
  \acro{DoS}{Denial-of-Service}
  \acro{FGSM}{Fast Gradient Sign Method}
  \acro{PGD}{Projected Gradient Descent}
  \acro{MLP}{Multilayer Perceptron}
  \acro{IP}{Intellectual Property}
  \acro{NLP}{Natural Language Processing} 
  \acro{PTQ}{Post Training Quantization} 
  \acro{QAT}{Quantization Aware Training} 
  \acro{RS}{Reverse Sigmoid} 
  \acro{AM}{Adaptive Misinformation} 
  \acro{DCP}{Deception} 
  \acro{CDCP}{Combined Deception}
  \acro{Soft-RS}{Softplus Reverse Sigmoid}
  \acro{OOD}{Out-of-Distribution}
  \acro{SGD}{Stochastic Gradient Descent}
  \acro{SVMs}{Support Vector Machines}
  \acro{SVM}{Support Vector Machine}
  \acro{LR}{Logistic Regression}
  \acro{LRs}{Logistic Regressions}
  \acro{DivQAT}{Divergence-Based Quantization Aware Training}
  \acro{IP}{Intellectual Property}
\end{acronym}

%% file: src/0-main.tex
% Main document with technical content

\section{Introduction}
\label{sec:sota}

    % \textit{1 to 1 1/2 pages \\
    % \textit{1~2 paragraphs per entry} \\
    % general context \\
    % specific context \\
    % general issues \\
    % specific issues \\
    % possible solutions \\
    % solutions limitations \\
    % your contribution \\
    % results \\
    % paper organization \\}

\input{src/introduction}

% \rewrite{Still needs more work}

\section{Background and related work}
\label{sec:sota}
\IEEEpubidadjcol
% \label{sec:basics}
    % \textit{1 page\\  
    % intro to general field of knowledge\\  
    % techniques\\  
    % state-of-the-art\\  
    %     table at the end, comparing with yours\\
    % intro to the next, highlighting the differences between yours and SOTA -> pointing to the improvements present in the next section\\}
    
\input{src/related-work}
% \rewrite{Still needs more work}

\section{Methodology}
\label{sec:methodology}

    % \textit{2-3 pages\\
    % introduction\\
    %     general idea for the solution\\
    %     recap the limitations on the SOTA\\
    %     small summary of how your thing goes\\}

\input{src/methodology}

\section{Experiments}
\label{sec:experiments}
\input{src/experiments}

    % 1-2 pages\\  
    % introduction\\  
    % results obtaining methodology\\  
    % results\\  
    % discuss\\  
    % (constraints - why)\\  
    % (possible limitations - workarounds)\\
% \section{Results}
\input{src/results}
\label{sec:results}

\section{Conclusion}
\label{sec:conclusion}

    % \textit{1-2 sentences per entry - past tense}\\
    % general context\\
    % specific context\\
    % general issues \\
    % specific issues\\
    % possible solutions\\
    % solutions limitations\\
    % your contribution\\
    % results\\

\input{src/conclusion}

%% file: src/introduction.tex
% **Introduction:**

% \kacem{change narrative to make less as a defense but more as a new training approach to make robustness by design}

In modern computing, the development of \ac{DL} models has resulted in a new era of innovation and technological advancement. These models, trained on vast datasets, have demonstrated remarkable capabilities across many tasks, from image recognition to natural language processing~\cite{krizhevsky2017imagenet,achiam2023-gpt4}. As the demand for intelligent systems continues to grow, there is a pressing need for deploying \ac{DL} models on edge devices, such as smartphones, and IoT devices, to enable real-time inference and decision-making in embedded systems.

Deploying \ac{DL} models on edge devices presents unique challenges, particularly in terms of computational resources, memory constraints, and energy efficiency. 
To address these challenges, researchers propose to compress \ac{DL} models without significantly affecting their performance using techniques such as \emph{pruning} and \emph{quantization}:
the former relies on reducing redundant and non-informative neurons~\cite{srinivas2015data, han2015learning}, and the latter reduces the number of bits required to represent the model weights \cite{vanhoucke2011improving, han2015deep}.
Quantization generally outperforms pruning due to its predictable memory savings and efficiency~\cite{kuzmin2024pruning}.
% Among several techniques, quantization has emerged as a widely used compression mechanism~\cite{cheng2017survey}. 
By reducing the precision of model parameters, quantization techniques effectively lower the computational costs and memory footprint of \ac{DL} models, making them more suitable for deployment on edge devices~\cite{survey_q}. 

\begin{figure}[t]
\centering
\includegraphics[width=0.45\textwidth]{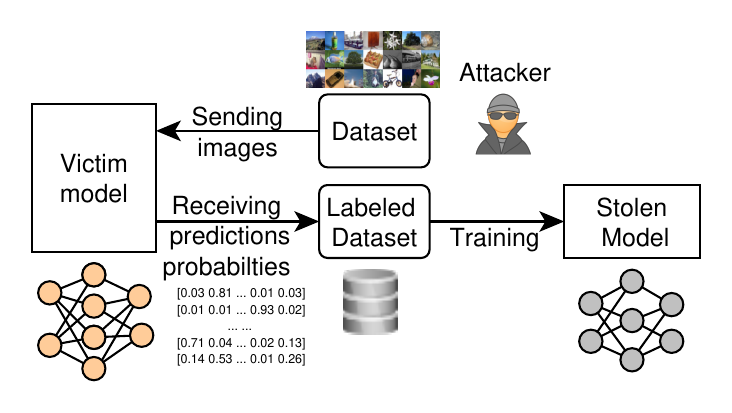}
\caption{In model extraction attacks, the adversary leverages the prediction probabilities of their inputs to build a labeled dataset. Then the attacker uses this dataset to train their stolen model.}
\label{fig:stealing-attack}
\end{figure}

\IEEEpubidadjcol

The widespread adoption of \ac{DL} techniques has raised concerns about their security and privacy~\cite{Goodfellow2018darkarts}. 
Particularly, in model extraction attacks, also called as model stealing attacks, adversaries can exploit the output predictions of a deployed model to extract a surrogate model. 
Fig.~\ref{fig:stealing-attack} illustrates the process of an extraction attack to steal a vision model. 
Even in a black-box setting, by observing the input-output pairs, the attacker can build a labeled dataset to train a surrogate \ac{CNN} with high accuracy and fidelity to the victim, posing a significant risk of \ac{IP} theft~\cite{tramer2016stealing}. 
% \kacem{
Extraction attacks are less studied in the literature than adversarial attacks since the latter have a direct impact on deteriorating the model's performance. 
Adversarial attacks focus on crafting adversarial examples to evade the classification or fool a model~\cite{goodfellow2014explaining}. 
However, extraction attacks are not of less importance since they may serve as a stepping stone to adversarial attacks and make them easier. An adversary could leverage the extracted model as a substitute victim to craft adversarial examples transferable to the original victim~\cite{papernot2017practical}. Hence, defending against extraction attacks makes other attacks more difficult.
% }

% \kacem{
State-of-the-art defenses against model extraction attacks still have multiple limitations.
Firstly, they do not modify the training algorithm to integrate the defense into training. 
In contrast to adversarial training~\cite{Bai_adversarial_training}, which was introduced to create models robust by design against adversarial examples, defenses against extraction attacks are still added as an afterthought to the models.
% }
% \kacem{State-of-the-art} 
The key idea of state-of-the-art defenses against model extraction attacks on computer vision models is often to inject noise into prediction probabilities~\cite{Orekondy2019poisoning,mazeika2022steer,kariyappa2020defending,lee2018defending,khaled2023}.
While effective to some extent, some of these defenses are computationally expensive and may not be feasible for real-time and edge implementations due to resource constraints and induced significant delays in inference time~\cite{Orekondy2019poisoning,mazeika2022steer}.
This prevents their integration into quantized models.
Other defenses preserve the model inference time and may be applied to quantized models without additional training~\cite{lee2018defending,khaled2023}. 
Yet, none of them have been customized for quantized models~\cite{lee2018defending,khaled2023}, nor do they focus on incorporating security into the training and design of quantized \ac{CNNs}.

% \IEEEpubidadjcol

In this paper, we introduce \emph{\ac{DivQAT}}, a novel approach that addresses the vulnerabilities of quantized models to model extraction attacks. \ac{DivQAT} is a technique designed to enhance the robustness of quantized models against stealing attacks on \ac{CNNs}. Unlike previous methods that add defenses as an afterthought, DivQAT integrates a model extraction defense directly into the \acf{QAT} process.
Our approach introduces a divergence loss term during \ac{QAT}, increasing the dissimilarity between the prediction probability distributions of the original and quantized models. 
% \kacem{
During quantization, instead of minimizing the cross-entropy loss of the quantized model with relation to the true labels, we additionally maximize the KL-divergence loss between the quantized and the original model predictions.
KL-divergence can be used to maximize the similarity between two models~\cite{hinton2015distilling}, thus our rationale, is that maximizing this loss would help deviate probabilities of the quantized model from the same distribution as the large one.
% }
By manipulating this divergence, we slightly poison prediction probabilities that are crucial for adversaries attempting model extraction, thus strengthening the robustness of quantized models without compromising their accuracy.  
% Furthermore, our induced perturbations do not compromise their accuracy. 

Through extensive experimentation, we demonstrate the effectiveness of our technique in enhancing the robustness of quantized models against extraction attacks, 
while preserving the accuracy.
% while also improving the performance of other defense mechanisms. 
Besides, we combine DivQAT with efficient state-of-the-art defenses that can be applied in quantized models and find that our training approach improves their performance.
% \mouna{while  preserving the accuracy}. 
% \mouna{make the other idea about the other defenses in a separate sentence: Besides, we combine our quantization-aware training approach with existing afterthought/post-training defenses [ ,,] and find that our novel training approach improves the performance of these other defense mechanisms. }
To summarize, we make the following contributions:
\begin{itemize}
    \item we propose a novel approach to QAT, which aims to enhance the robustness of quantized models against model extraction attacks.  To the best of our knowledge, our technique DivQAT is the first to modify the quantization process to integrate a model extraction defense directly into the training phase;
\item through empirical validation on benchmark vision datasets and against state-of-the-art attacks, we demonstrate the efficacy of our technique in mitigating the risks posed by model extraction attacks, without compromising the accuracy of the model, and;
\item when combined with other post-training defense mechanisms, our quantization technique, DivQAT, improves their effectiveness compared to traditional QAT, providing an additional layer of security for edge devices deploying quantized models.
\end{itemize}

In the remainder of this paper, we provide an overview of the background and related work in Section~\ref{sec:sota}, 
we delve into the methodology behind our proposed divergence-based quantization technique in Section~\ref{sec:methodology}, 
we present and discuss the experimental results in Section~\ref{sec:experiments}, and finally, we conclude the paper in Section~\ref{sec:conclusion}. 

% The remainder of the paper is organized as follows: 
% Section~\ref{sec:ML-DL} introduces the basic notions to \ac{ML} and \ac{DL}; 
% Section~\ref{sec:background-sota} reviews the background and the state-of-the-art that relates to privacy and security threats and attacks; 
% Section~\ref{sec:methodology} details our methodology; 
% we include our experiments and obtained results in Section~\ref{sec:experiments}; 
% Section~\ref{sec:defenses} summarizes potential defenses and countermeasures against extraction attacks;
% % we review defenses to extraction attacks in Section~\ref{sec:defenses};
% and Section~\ref{sec:conclusion} concludes this paper.

%% file: src/related-work.tex
In this section, we introduce the background and basic concepts to \ac{ML} and the quantization of \ac{CNNs} mentioned in our work. Then, we review the related work to model extraction attacks and defenses. 

\subsection{\acf{ML}}

\emph{\ac{ML}} algorithms are computational methods that learn from data to perform specific tasks~\cite{Goodfellow-et-al-2016}. This data comprises examples with quantitatively measured features, each represented as a vector in a multi-dimensional space. 
% ML algorithms are broadly categorized as supervised and unsupervised~\cite{Goodfellow-et-al-2016}.
% \emph{Supervised learning} algorithms work with annotated datasets, meaning each example in the dataset is associated with a provided label. The algorithm learns to predict the label from the example by estimating the probability of the label given the example.
% \emph{Unsupervised learning} algorithms handle an unlabeled dataset containing a collection of examples and try to learn interesting properties about the dataset’s structure. The algorithm aims to learn the probability distribution that generated the data~\cite{Goodfellow-et-al-2016}.
% \emph{Semi-supervised learning} falls between supervised and unsupervised learning, utilizing both labeled and unlabeled data to improve learning accuracy. 
% \emph{Self-supervised learning}, on the other hand, generates its own labels from the data, often through pretext tasks, to learn representations without explicit manual labeling~\cite{zhai2019s4l}.
% In addition to these, there are other \ac{ML} algorithms that interact with a dynamic environment, such as \emph{reinforcement learning} algorithms. In reinforcement learning, the learning system, or agent, learns a policy that maximizes a reward signal through interaction with the environment~\cite{Russell2003}.

\emph{\ac{DL}}, a subset of \ac{ML}, involves techniques based on Artificial Neural Networks (ANNs)~\cite{Goodfellow-et-al-2016}. \ac{DL} models map input vectors to output values through multiple layers of interconnected neurons. These models are trained by optimizing their parameters to minimize a loss function, typically using methods like \ac{SGD}~\cite{Goodfellow-et-al-2016}.
SGD is an iterative optimization method commonly used for minimizing a cost function in ML models. During training, SGD processes one training example (or a small batch) at a time. It updates the model parameters by taking a step in the direction of the negative gradient of the cost function. The learning rate controls the step size during parameter updates. The loss function (often denoted as $\mathcal{L}$) quantifies the difference between the predicted values and the actual labels. For instance, the cross-entropy loss equation is defined as:
\begin{equation}
    \label{eq:loss_ce}
     % L(y,f_{W_{Q}}) = -\sum_{i}y_{i}\log f_{W_{Q}}(x_{i})  
     % \mathcal{XE}(y,f_{W_{Q}}(x)) = -\sum_{i}y_{i}\log f_{W_{Q}}(x_{i})  
     \mathcal{L}_{CE}(y, f_{W}(x)) = -\sum_{i} y_i \log f_{W}(x_i)
    % \min_{W_{Q}} L(y,f_{W_{Q}}) = \min_{W_{Q}} \mathcal{XE} (y_{i};f_{W_{Q}}(x_{i}))  
\end{equation}
where $y_i$ is the actual label for the $i$-th example, $f_{W}(x_i)$ is the predicted value (i.e., the output of the model) for the $i$-th example and $W$ represents the model parameters.
In each iteration of SGD, the gradient of the loss function with respect to the model parameters is computed, and the parameters are updated accordingly. The process repeats for a specified number of iterations or until convergence. 
% \textcolor{cyan}{explain more the loss function, cite and add examples, it relates to your main contribution}
% \kacem{DONE}

% \emph{Generative models}, a class of \ac{ML} algorithms, aim to learn the true data distribution of the training set so as to generate new data points. These models, including Generative Adversarial Networks (GANs), are widely used in unsupervised learning~\cite{goodfellow2020generative}
\emph{Generative models} are a class of \ac{ML} models designed to create new data samples that resemble a given dataset. They learn the underlying patterns and generate novel instances. Popular examples include Variational Autoencoders (VAEs)~\cite{kingma2019introduction} and Generative Adversarial Networks (GANs)~\cite{goodfellow2020generative}. 
% \textcolor{cyan}{the part about generative models is irrelevant}

% \emph{Transformers} and \emph{Large Language Models (LLMs)} are recent innovations in deep learning that have significantly advanced the field of natural language processing. Transformers utilize self-attention mechanisms to process sequences of data efficiently, allowing for more parallelization and better handling of long-range dependencies~\cite{vaswani2017attention}. Large language models, such as GPT-3 and GPT-4, are built upon transformer architectures and trained on extensive corpora, enabling them to generate coherent and contextually relevant text~\cite{brown2020language-gpt3,achiam2023-gpt4}. These models have set new benchmarks in a variety of language tasks, from translation to content creation, demonstrating an unprecedented understanding of human language. 
% \textcolor{cyan}{this paragraph is irrelavant, you do not use LMMS or even text, }

\emph{\acf{CNNs}} are a specific type of \ac{DL} models that are particularly suitable for tasks involving images. The input to a CNN is typically an image, and the features are the individual pixels of the image. \ac{CNNs} use convolution operations and non-linear activations to learn filters and characteristics relevant to the task~\cite{Krizhevsky2017}. These filters can capture local patterns, like edges and shapes, in the input image. After sufficient training, CNNs can learn high-level features in images that are useful for tasks such as image classification. This makes CNNs highly effective for computer vision tasks. 
% \textcolor{cyan}{this paragraph should follow the DL paragraph directly,  }

\subsection{Quantization of \ac{CNNs}}
% \textcolor{cyan}{put the rest of subsection ML in the begining of this subsection}

Quantization is a fundamental technique used to reduce the computational and memory requirements of \ac{DL} models, particularly when deploying them on resource-constrained devices such as edge devices~\cite{han2015deep,vanhoucke2011improving}. In the context of Convolutional Neural Networks (CNNs), quantization involves reducing the precision of the model parameters, such as weights and activations, from floating-point numbers to lower bit-width integers~\cite{Wu_2016_CVPR}. 
Quantization techniques fall into two families: \acf{PTQ} and \acf{QAT}~\cite{survey_q}.
% \textcolor{cyan}{are there other types of quantization that do not reduce the precision?}

% Post-Training Quantization (PTQ) 
\ac{PTQ} is a commonly used approach where quantization is applied after the model has been trained using full precision (usually 32-bit floating-point numbers). In PTQ, the model is first trained using standard techniques, and then the weights and activations are quantized to lower precision (e.g., 8-bit integers) during inference. While PTQ is straightforward to implement and is effective in reducing model size and inference latency, it may suffer from accuracy degradation due to quantization-induced errors~\cite{survey_q}.

% Quantization Aware Training (QAT)
\ac{QAT}, on the other hand, integrates the quantization process into the training pipeline. During QAT, the model is trained with quantization in mind, simulating the lower precision arithmetic that will be used during inference~\cite{jacob2018quantization}.
This allows the model to adapt to the quantization effects, often resulting in a quantized model that closely approximates the accuracy of the original full-precision model. QAT involves additional training for a pre-trained model with quantization constraints, which requires additional computational resources and time compared to PTQ but can yield superior performance on quantized models~\cite{survey_q}.
% \textcolor{cyan}{the last sentence is only valid for pretrained models?}

% % https://leimao.github.io/blog/PyTorch-Static-Quantization/
% % https://leimao.github.io/blog/PyTorch-Quantization-Aware-Training/

\subsection{Model stealing attacks}

% - API-based attacks

% - side-channel attacks

% - how they differ from knowledge distillation

Model extraction attacks on \ac{DL} models, also known as model stealing attacks, pose a significant threat to the \ac{IP} of \ac{ML} model owners. 
These attacks occur during the inference phase, even in black-box settings, where an adversary can observe the prediction outputs of their queries to a model and attempt to replicate its functionality or parameters~\cite{tramer2016stealing}. 
The goal is to create a classifier that matches or closely approximates the target classifier, which not only risks the IP but also facilitates other attacks like membership inference~\cite{shokri2017membership} and evasion attacks~\cite{papernot2016crafting}.

Model stealing attacks are categorized into two types: side-channel extraction attacks (leveraging hardware side-channel information)~\cite{Wei2018sca,Hu2020sca,Wei2020sca,Xiang2020sca} and API-based extraction attacks (exploiting input-output predictions through APIs)~\cite{tramer2016stealing, papernot2017practical,orekondy2019knockoff,Krishna2019,Reith2019, Truong2020, Pal2020,Yuan2020,Jagielski2020,kariyappa2021maze,truong2021data,Zhou_2020_CVPR_dast,Sanyal_2022_CVPR_hard}.
In this paper, we focus on stealing attacks that leverage API access.

Earlier work~\cite{papernot2017practical} introduces a method to generate synthetic data using Jacobian-based data augmentation techniques. Then using this data they query the target model. They use the stolen model to craft adversarial examples~\cite{szegedy2013intriguing} 
to evade the classification of the victim model.
An intriguing attack, known as \emph{KnockoffNets}~\cite{orekondy2019knockoff}, aims to create a functionally equivalent clone of \ac{CNNs} using random and publicly available data. Despite its simplicity, KnockoffNets outperforms other state-of-the-art stealing attacks.

Recent advancements in extraction attacks against \ac{CNNs} include \emph{MAZE}~\cite{kariyappa2021maze} and \emph{DFME}~\cite{truong2021data}. These approaches leverage generative models for building the adversary’s dataset, without relying on public datasets. They generate query images while observing the output predictions of the victim model to target the decision boundary better. They achieve successful extraction for most cases, albeit at the cost of requiring millions of queries compared to KnockoffNets.

Other stealing attacks in the literature assume that a target model can only output 
top-1 predictions (hard labels) instead of probability values~\cite{Zhou_2020_CVPR_dast,Sanyal_2022_CVPR_hard}. This could limit the amount of information that can be obtained about the model’s confidence in its predictions.
In our work, we only consider attacks leveraging soft labels with full access to output probabilities.
% \mouna{what about refrences 6, 26,27,28,29,30,et 31 ? }

\subsection{Model stealing defenses}

% - perturbation based defenses

% - defenses based on obfuscation

% - poisoning attacks

% - knowledge distillation based defenses and how they are different (nasty teacher)
% \\

% To enable  model owners to claim their models if stolen, previous work proposes watermarking the model \cite{jia2020entangled,szyller2019dawn}. 
% Watermarking can be done by changing the output probabilities for a small subset of queries. However, this technique does not prevent model stealing.

% \mouna{this is very long paragraph, please re-struture it to be more clear}
% Defense approaches in the literature, propose to 
To protect models against API-based model stealing attacks, researchers propose to add perturbation to the prediction probabilities~\cite {lee2018defending,kariyappa2020defending,Orekondy2019poisoning,mazeika2022steer,khaled2023}.
Since the adversary relies on these predictions to build a labeled dataset, the defender aims to poison the attacker's training data with perturbed annotations.

% Defense approaches in the literature, 
% aiming to prevent API-based model stealing attacks,
% propose to 
% protect models by perturbing the prediction probabilities~\cite {lee2018defending,kariyappa2020defending,Orekondy2019poisoning,mazeika2022steer,khaled2023}. 

The \emph{\ac{RS}} defense \cite{lee2018defending} proposes to change the activation function of the last layer of a \ac{NN} to a reverse sigmoid to perturb the output predictions. However, the transparency of the model is negatively affected by using such a high perturbation level in this technique.
% These perturbation-based defenses can be categorized into two categories: \emph{proactive} and \emph{reactive} techniques.  
Another method includes truncating the probability scores~\cite{orekondy2019knockoff} but it is ineffective against state-of-the-art attackers due to its simplicity.
Additionally, adding random noise is explored~\cite{orekondy2019knockoff} but is considered a bad strategy as it compromises the defender's performance due to large perturbation levels.
% Other methods also rely on injecting high perturbations by adding random noise~\cite{orekondy2019knockoff}. But, they are ineffective against state-of-the-art attackers because they compromise the defender's performance.
Some defenses propose monitoring the input queries to detect an ongoing attack by identifying a deviation from the normal distribution~\cite{Juuti2019} or a high volume of \ac{OOD} queries~\cite{kariyappa2020defending}. 
In \emph{\ac{AM}}~\cite{kariyappa2020defending} defense, 
queries with low prediction scores are marked as \ac{OOD}, and using an auxiliary model, a high noise is injected into the predictions.
% \mouna{explain + link this to the previous sentence , about Out Of Distribution queries?}
Nevertheless, this method is susceptible to false positives, which deteriorates the model's performance.
% \mouna{why they are ineffective?  + why you cite knockoffs? }
% Authors in~\cite{ma2021undistillable} propose \emph{Nasty Teacher} aiming to maximize the deviation between the output of their model and a normal pre-trained one. However, this defense focuses only on knowledge distillation cases, where knowledge is transferred from pre-trained teacher models to student models~\cite{hinton2015distilling}. They do not explore state-of-the-art model stealing attacks which are larger and more practical threats.

Recent works~\cite{Orekondy2019poisoning,mazeika2022steer} propose to compute the appropriate perturbation level per query relying on optimization methods.
\emph{MAD}~\cite{Orekondy2019poisoning} is a defense that attempts to solve an optimization problem where it mimics the attacker by using a different network and tries to add the perturbation that maximizes the error of the simulated surrogate model. It aims to deviate the adversary's gradient signal by injecting targeted noise into the prediction probabilities.
However, computing these perturbations is computationally expensive and delays the inference latency. 
% \mouna{what's the limitations of MAD?}
\emph{GRAD$^2$}\cite{mazeika2022steer} is a method similar to MAD, focusing on redirecting the gradient signal. It suggests simulating the attacker's knowledge by training a surrogate model using the attacker's queries. However, GRAD$^2$ assumes that the attacker will send a large batch of queries for training the simulated attacker, which limits its practicality. This is because it presumes that the system where the model is deployed for predictions has enough hardware resources for the training. 

\emph{\ac{DCP}}~\cite{khaled2023} is a heuristic defense that perturbs output probabilities without additional training nor relying on an auxiliary model. Their technique is inspired by the defenses RS and AM, but it achieves better trade-offs between the defender's accuracy and the perturbation level while maintaining low inference latency. The authors demonstrated the effectiveness of their defense on large \ac{CNNs} and their quantized counterparts.
However, DCP does not modify the quantization process but rather uses a simple heuristic function to compute the noise that can be injected at the last layer of a model during inference. 
To summarize, current strategies to defend against model extraction attacks on computer vision models typically involve adding noise to predictions. However, none of these defenses~\cite{Orekondy2019poisoning,mazeika2022steer,kariyappa2020defending,lee2018defending,khaled2023} are tailored for quantized models, nor do they explore integrating security into the training and design of quantized \ac{CNNs} (Table~\ref{tab:compare}).
Additionally, 
some of these strategies~\cite{Orekondy2019poisoning,mazeika2022steer} are computationally demanding and may not be suitable for real-time and edge deployments due to resource limitations and significant inference time delays.
% }

\begin{table}[t]
\centering
\caption{Our approach Vs. state-of-the art methods defending against extraction attacks}
\label{tab:compare}
% \begin{tabular}{|p{2cm}|p{2cm}p{2cm}p{2cm}p{2cm}|}
% \begin{tabular}{p{2cm}p{2cm}p{3cm}p{2.5cm}}
\begin{tabular}{ccccc}
% \begin{tabular}{|p{1.4cm}|p{1.2cm}|p{2.3cm}|p{1.cm}|p{0.7cm}|}
% \hline
\toprule
Approach & {Fast} & {No auxiliary} & {Low} & {Integrated} \\ 
 & {inference} & {model} & {perturbation} & {in training}\\ % & {Accuracy preserving} \\
% \hline
\midrule
MAD \cite{Orekondy2019poisoning}& \xmark & \xmark & \cmark  & \xmark  \\ % & \cmark   \\
GRAD$^2$ \cite{mazeika2022steer}   & \xmark & \xmark & \cmark  & \xmark  \\ % & \cmark  \\
AM \cite{kariyappa2020defending}& \cmark & \xmark   & \cmark  & \xmark  \\ % & \cmark   \\
RS \cite{lee2018defending}      & \cmark & \cmark  & \xmark  & \xmark  \\ %  &\cmark   \\
DCP \cite{khaled2023}  & \cmark & \cmark & \cmark  & \xmark  \\ % & \cmark \\
\textbf{DivQAT}   & \cmark & \cmark & \cmark & \cmark  \\ % & \cmark \\
% \hline
\bottomrule
\end{tabular}
\end{table}

%% file: src/methodology.tex
In this section, we present the threat model assumptions. Then, we describe our approach DivQAT, the validation process in our methodology, and the evaluation metrics.

\begin{figure*}[!t]
\centering
\includegraphics[width=0.95\textwidth]{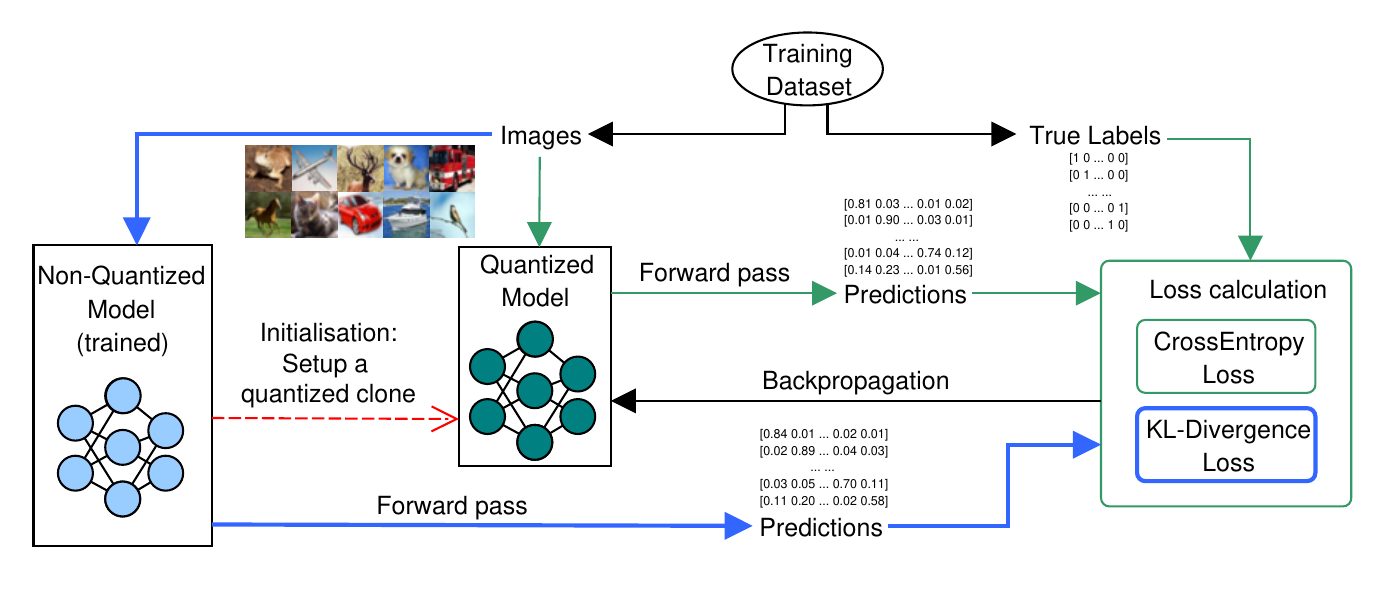}
\caption{Our Div-QAT approach: First, we initialize a quantized neural network based on a trained non-quantized model (red dashed arrow). Then, the quantized model is trained on the same training dataset as the original model. In each training step, input images are fed to the quantized model (forward pass) to provide prediction probabilities which will be used along with the true labels to compute the cross-entropy loss (green arrows). Additionally, we feed the training examples to the original (non-quantized) model and get their predictions to compute the KL-divergence between them and the quantized model's predictions (blue bold arrows). After that, the loss (Eq.~\ref{eq:loss_div_qat}) is computed. The quantized neural network weights are updated through backpropagation and the training process continues. }
\label{fig:approach}
\end{figure*}

\subsection{Threat model}
% We assume that the attacker is only granted an API access to the victim model and his behavior is similar to a benign user.
% The attacker cannot access the victim model's architecture, nor its training data or parameters.
% Model stealing attacks work even in this black-box setting where the user can only observe the input-output pairs. 
% Through leveraging the victim's predictions on its input, the attacker builds a labeled dataset and uses it to train their model.
% The attack seeks to create a duplicate model with a close functionality to the victim.
% \kacem{
After quantization, the model owner (i.e., the victim) provides an API access to users where they can send inputs (e.g., images) and receive output probabilities (e.g., 90\% being a vehicle).
% }
We consider a scenario where the attacker has API access to the victim's quantized model, akin to a benign user's interaction. 
The attacker lacks direct access to the model's architecture, training data, or parameters. 
Despite these limitations, model stealing attacks remain feasible through the observation of input-output pairs~\cite{orekondy2019knockoff}. 
The attacker's objective is to replicate the victim model's functionality closely. 

\subsection{Approach}
% Most API-based model stealing attacks leverage the output probabilities. We intend to deviate these predictions from their standard distribution to make the attack less successful.
% Specifically, we modify the quantization-aware training process to obtain a more robust quantized model. 
% Our goal is to create a model that generates slightly divergent probabilities compared to the non-quantized model, while still preserving the same level of accuracy.
% Our rationale is that when we add divergence to the probabilities, we poison the dataset constructed by the attacker, making it harder to steal the model.

Our approach focuses on disrupting the output probabilities that API-based model stealing attacks typically exploit. We propose an alteration to the QAT process, aiming to produce a quantized model that, while maintaining accuracy, yields probabilities divergent from the original model. 
We introduce \emph{DivQAT}: a divergence-based \ac{QAT} algorithm. This divergence introduces noise into the attacker's dataset, hindering their ability to replicate the model effectively.

% \kacem{
We implement our approach by integrating a divergence-based loss term into the QAT process. 
Figure~\ref{fig:approach} describes the steps of our algorithm.
% }
% During each training epoch, inputs are fed to both the original and quantized models to obtain their respective predictions. The divergence is then computed and used to adjust the quantized model's training trajectory
To illustrate, let $f_{W_{L}}$ represent a large (i.e., non-quantized) model trained with a dataset $\mathcal{X}$ and  $f_{W_{Q}}$ denote its quantized counterpart, with \( W_L \) and \( W_Q \) being their respective parameters. Let $x_i$ and $y_i$ denote a training example and its ground truth label. In a normal quantization process, the quantized network is trained by minimizing cross-entropy loss, defined as:

% To illustrate, let \( f_{W_L} \) represent the original, non-quantized model trained on dataset \( \mathcal{X} \), and \( f_{W_Q} \) denote its quantized counterpart, with \( W_L \) and \( W_Q \) being their respective parameters. For a given training example \( (x_i, y_i) \), the conventional QAT minimizes the cross-entropy loss:

\begin{equation}
    \label{eq:loss_qat}
     % L(y,f_{W_{Q}}) = -\sum_{i}y_{i}\log f_{W_{Q}}(x_{i})  
     % \mathcal{XE}(y,f_{W_{Q}}(x)) = -\sum_{i}y_{i}\log f_{W_{Q}}(x_{i})  
     \mathcal{L}_{CE}(y, f_{W_Q}(x)) = -\sum_{i} y_i \log f_{W_Q}(x_i)
    % \min_{W_{Q}} L(y,f_{W_{Q}}) = \min_{W_{Q}} \mathcal{XE} (y_{i};f_{W_{Q}}(x_{i}))  
\end{equation}

% \textcolor{cyan}{the second term isFWQ or FWL?}
% \kacem{
Our proposed methodology modifies the loss function by incorporating a divergence term, specifically the Kullback-Leibler (KL) divergence~\cite{kullback1951information}.
We modify the loss to minimize the cross-entropy loss of the quantized model and to maximize the KL-divergence between the large and the quantized model. 
Contrary to previous work that used KL-divergence as a mean to maximize the similarity between two models while transferring the knowledge between them~\cite{hinton2015distilling,kariyappa2021maze}, 
our rationale is that using this loss would help deviate probabilities of the quantized model from the same distribution as the large one. 
For example, in knowledge distillation~\cite{hinton2015distilling}, where knowledge is transferred from a large network to a smaller one, both the cross-entropy loss and the KL-divergence are minimized. Furthermore,  extraction attacks such as MAZE~\cite{kariyappa2021maze} leverage minimizing the KL-divergence while training the stolen model, to have a surrogate model that closely mimics the victim.
% }

% \kacem{
In our DivQAT algorithm, we seek to have a quantized model that does not imitate the decision boundary of the original model but rather have a more generalized one that makes it harder to extract knowledge from. 
During quantization, when computing the KL-divergence loss in each training epoch, 
% in addition to the quantized model predictions
we additionally feed the training examples to the large model and get their predictions.
The large model and the quantized model predictions are respectively denoted as $f_{W_{L}}(x_{i})$ and $f_{W_{Q}}(x_{i})$ for an input $x_i$. 
% Figure~\ref{fig:approach} demonstrates our technique.
% When computing the loss in each training epoch, we feed the training examples to the large model as well as the quantized model and get their predictions denoted as $f_{W_{L}}(x_{i})$ and $f_{W_{L}}(x_{i})$ for an input $x_i$. 
% The KL-Divergence is computed as:
% }
% Our proposed methodology modifies this loss function by incorporating a divergence term, specifically the Kullback-Leibler (KL) divergence~\cite{kullback1951information}, between the predictions of \( f_{W_L} \) and \( f_{W_Q} \). 
Then, the KL-divergence is calculated as follows:

\begin{equation}
    \label{eq:loss_kl} 
     % \mathcal{KL}(f_{W_{Q}}(x),f_{W_{L}}(x)) = \sum_{i}f_{W_{Q}}(x_{i})\log \frac{f_{W_{Q}}(x_{i})}{f_{W_{L}}(x_{i})}
     D_{KL}(f_{W_Q}(x) \parallel f_{W_L}(x)) = \sum_{i} f_{W_Q}(x_i) \log \frac{f_{W_Q}(x_i)}{f_{W_L}(x_i)}
\end{equation}

% Therefore, the loss function of our algorithm is defined as:
Consequently, our proposed loss function becomes:

\begin{equation}
\begin{split}
    \label{eq:loss_div_qat}
%      L(y,f_{W_{Q}},f_{W_{L}}) = \mathcal{XE}(y,f_{W_{Q}}(x)) -
% \alpha  \mathcal{KL}(f_{W_{Q}}(x),f_{W_{L}}(x))
\mathcal{L}(y, f_{W_Q}, f_{W_L}) = &\mathcal{L}_{CE}(y, f_{W_Q}(x)) \\
&- \alpha D_{KL}(f_{W_Q}(x) \parallel f_{W_L}(x))
\end{split}
\end{equation}
Here, \( \alpha \) is a hyperparameter that we add to balance between the cross-entropy loss and the KL-Divergence loss and hence it controls the trade-off between accuracy and robustness against model extraction. The training relies on minimizing the loss function, thus a negative KL-divergence term would push the model to maximize the divergence between the probability distributions while minimizing the cross-entropy loss to maintain the accuracy.

% the trade-off between accuracy and robustness against model extraction.

% \subsection{Implementation}

% After the training, we perform several model stealing attacks on the quantized model. \textcolor{cyan}{this sentence goes to the next paragraph}

% The attacker has API access to the quantized model to which he sends queries $x'$ to obtain predictions $y'=f_{W_{Q}}(x')$. 
% Then, he uses this labeled dataset to train a stolen model $f_{W_{S}}$ minimizing the cross-entropy loss, defined as:
% \begin{equation}
%     \label{eq:loss_adv}
%      L(y',f_{W_{S}}(x')) = -\sum_{i}y'_{i}\log M_{S}(x')_{i}  
% \end{equation}

\subsection{Validation}
% \kacem{
Validating our approach, we seek to answer multiple research questions:
\begin{itemize}
    \item Does our quantization algorithm enhance the robustness of quantized models against extraction attack?
    \item How does the robustness of models obtained through DivQAT compare to other quantized models and baseline defenses applicable to quantized models?
    \item What is the impact of combining our quantization algorithm with other defenses?
    \item Does DivQAT work on different quantization configurations?
\end{itemize}
% }

We use several model stealing attacks to validate the robustness of quantized \ac{CNNs} obtained by our algorithm \emph{DivQAT}.
We perform traditional stealing attacks that rely on publicly available datasets to query the model namely KnockoffNets~\cite{orekondy2019knockoff}. Additionally, we leverage data-free stealing attacks that utilize generative models to synthesize data, namely DFME~\cite{truong2021data} and MAZE~\cite{kariyappa2021maze}.

We compare the robustness of models obtained through \emph{DivQAT} with
the original \emph{large} model, quantized models obtained through \emph{QAT} and \emph{PTQ} as well as other defenses. 
To the best of our knowledge, the state-of-the-art defenses that are suitable for edge device implementations on quantized models are: 
\emph{\ac{DCP}}~\cite{khaled2023}, and \emph{\ac{RS}}~\cite{lee2018defending}. 
These defenses maintain the inference time without requiring additional models.

Then, we evaluate the impact of our quantization algorithm on other post-training defenses against stealing attacks.
In this validation step, we train models using DivQAT and QAT. Then, after applying \ac{DCP} and \ac{RS} defenses to models obtained through both quantization algorithms, we evaluate their robustness against stealing attacks.

% \kacem{
Finally, we test our approach on different variations of \emph{QAT}. The quantization algorithm can be adapted depending on the target device which will be used during inference: \emph{mobile} or \emph{server}.
% On the other hand, during normal QAT, the quantized model is initialized based on the weights from the large model, and then during training, it is fine-tuned. Alternatively, we initialize the quantized model weights randomly and we train it from scratch.
We quantize models using both configurations, then we validate their robustness against extraction attacks.
% }

\subsection{Evaluation metrics}
The performance of defenses against stealing attacks is measured through the adversary model's prediction results on the victim's test data. 
To estimate how well a stolen adversary model performs on the same task as the victim, we follow previous work in this research area~\cite{mazeika2022steer,khaled2023,lee2018defending} and compute the \emph{Adversary's classification error}.
This metric measures the stolen functionality from the victim, i.e., the performance of the extracted model on the same task. 
Furthermore, to estimate the fidelity of the predictions of the adversary model to the victim, we compute the \emph{Disagreement} between them. 
This metric is obtained by calculating the fraction of mismatching predictions from the victim and the extracted model. This is similar to calculating the classification error of the extracted model using the predictions of the attacked victim as ground truth labels. A powerful adversary would have lower \emph{Adversary's classification error} and \emph{Disagreement}, hence we seek to maximize both metrics to weaken the adversary and hinder the model extraction attack.

To measure the impact of our defense algorithm on the victim model, similar to \cite{orekondy2019knockoff,mazeika2022steer,khaled2023}, we use two utility metrics: the \emph{Defender's classification error}  and the \emph{$\ell_1$ distance} between the clean predictions and the perturbed predictions by each defense technique.
The $\ell_1$ distance metric measures the magnitude of the perturbation that impacts the prediction probabilities, i.e., it is the distance between the clean probabilities and the perturbed ones. 
For example, an $\ell_1$ distance larger than $1$ entails a high perturbation that might harm the model's transparency and reduce its utility when probabilities are required to be precise~\cite{khaled2023}.

%% file: src/experiments.tex
% \subsection{Setup}

In this section, we explain our experimental setup including the datasets, the neural network architectures, and the training process.
Then, we show and analyze our experimental results validating our approach.
% % ----------------------

\subsection{Experimental setup}
\subsubsection{Victim datasets}

%%% DATASETS TABLE
\begin{table}[!t] %[htbp]
  \centering
  \caption{Selected datasets in our experiments}
%   \begin{tabular}{|p{1.5cm}|p{3.3cm}|p{1.8cm}|p{2.6cm}|p{2.3cm}|}
%   \begin{tabular}{|p{1.2cm}|p{2.7cm}|p{0.8cm}|p{1.2cm}|p{0.8cm}|}%p{2cm}|
  % \begin{tabular}{|p{1.4cm}|p{1.2cm}|p{2.3cm}|p{1.cm}|p{0.7cm}|}%p{2cm}|
  \begin{tabular}{>{\centering\arraybackslash}p{1.cm}>{\centering\arraybackslash}p{1.44cm}>{\centering\arraybackslash}p{2.35cm}>{\centering\arraybackslash}p{.7cm}>{\centering\arraybackslash}p{1.24cm}}%p{2cm}|
    % \hline
    \toprule
      \centering
    % \rowcolor[gray]{0.8}\color{black}
    \textbf{Dataset}    & \textbf{Dataset size}  &   \textbf{Description}   &  \textbf{Classes} & \textbf{Samples} %& \textbf{Image size}
    \\ 
    \cmidrule{1-5} %\hline
    % MNIST   &  Handwritten grayscale digits & 28$\times$28 & Train: 50k \newline   Test: 10k & 10 %& TODO   
    % \\ \hline
    CIFAR10   & Train: 60k  \newline Test: 10k  & Images of animals and vehicles  &  10 & 
    \multirow{2}{*}{\includegraphics[width=16px]{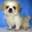}}
    \multirow{2}{*}{\includegraphics[width=16px]{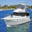}}  %& 32$\times$32
    \\ \cmidrule{1-5}%\hline
    CIFAR100   &  Train: 60k  \newline Test: 10k  & Diverse real-world images.   & 100 & 
    \multirow{2}{*}{\includegraphics[width=16px]{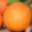}}
    \multirow{2}{*}{\includegraphics[width=16px]{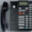}}  % &32$\times$32
    \\ \cmidrule{1-5}%\hline
    SVHN   & Train: 73k \newline Test: 26k  & Images of street view house numbers  &  10 & 
    \multirow{2}{*}{\includegraphics[width=16px]{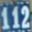}}
    \multirow{2}{*}{\includegraphics[width=16px]{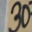}}% Train: 73257 Test: 26032 & 32$\times$32
    \\ \cmidrule{1-5}%\hline %& TODO
    % \hline
    % Fashion MNIST   &  Grayscale images of clothes & 28$\times$28 & Train: 50k \newline   Test: 10k & 10 %& TODO
    % \\\hline
    GTSRB   &   Train: 39k  \newline Test: 12k  & German traffic signs dataset  &  43  & 
    \multirow{2}{*}{\includegraphics[width=16px]{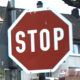}}
    \multirow{2}{*}{\includegraphics[width=16px]{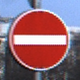}} %  &  32$\times$32 (resized)
    % \\ \hline
    % ImageNet  &  Various real-world images belonging to 1000 different categories & 224$\times$224 &  Train: 1.2M \newline Test: 50k  & 1000
    % \\\hline
    % CUB-200   &  Images of 200 bird species & 224x224 & 5994 / 5794  & 200 %& TODO   
    % \\\hline
    % Caltech-256  &  Various real-world images belonging to 256 different categories & 224x224 &  23703 / 6904  & 256 %& TODO   
    \\ \bottomrule %\hline
    
  \end{tabular}
  \label{tab:Datasets}
\end{table}

We leverage \ac{CNNs} trained on 4 benchmark vision datasets: CIFAR-10\cite{krizhevsky2009cifar10}, CIFAR-100~\cite{krizhevsky2009cifar10},  SVHN~\cite{netzer2011svhn} and GTSRB~\cite{Stallkamp2011gtsrb}. We describe these datasets in Table~\ref{tab:Datasets}. CIFAR-10 is an image dataset of 10 classes of animals and vehicles. CIFAR-100 contains 100 classes of diverse real-world images such as objects, people, animals, fruits, and vegetables. SVHN is a dataset of 10 classes of images of house numbers taken from street views. GTSRB is a dataset of 43 classes of german traffic signs.
We consider models trained on these datasets as potential victims of stealing attacks.

% \textcolor{cyan}{maybe call this paragraph 2)Stealing attacks design ? and start it with the sentence that says: we apply three stealinkg attacks: knockoffnets, maze and dfme.} 
\subsubsection{Adversary datasets}
We utilize three stealing attacks: KnockoffNets~\cite{orekondy2019knockoff}, DFME~\cite{truong2021data} and MAZE~\cite{kariyappa2021maze}.
An adversary that leverages the KnockoffNets~\cite{orekondy2019knockoff} attack requires a publicly available dataset to perform the attack.
In accordance to~\cite{mazeika2022steer,khaled2023}, we reproduce two attack scenarios in which (a) an attacker has access to images situated close to the data distribution of the victim, but without any similar data between them, and (b) an attacker who lacks knowledge of the data distribution, thus uses \ac{OOD} data for the attack.
Therefore, to perform the first one, since CIFAR-10 and CIFAR-100 are subsets of the same distribution which is the TinyImages dataset~\cite{krizhevsky2009cifar10}, 
we choose images from CIFAR-10 to attack models trained on CIFAR-100, and the opposite way around.
For adversaries with limited knowledge, we opt for unrelated datasets. CIFAR-10 images are selected as adversary datasets to attack victims trained on GTSRB and SVHN.
In data-free model stealing attacks DFME~\cite{truong2021data} and MAZE~\cite{kariyappa2021maze}, the attacker generates the adversary dataset during the attack while querying the victim model and observing its predictions.

\subsubsection{Models architectures and training}
We select the ResNet-18~\cite{he2016deep} architecture to train the non-quantized neural networks for $50$ epochs. 
Quantization (i.e., QAT and DivQAT) is performed with an SGD optimizer with a momentum of $0.9$ and a weight decay of $5\cdot10^{-4}$. The learning rate is initialized at $0.01$ and updated using a cosine annealing schedule. 
% \textcolor{cyan}{cite somehting for the QAT that also uses ResNet?}
% The target network architecture for the quantization process is 
In KnockoffNets attack~\cite{orekondy2019knockoff}, 
we select four different CNN architectures for the adversary model:
ResNet-18~\cite{he2016deep}, 
ResNet-34~\cite{he2016deep}, MobileNetV2~\cite{Sandler_2018_MobileNetV2}
% ResNeXt-29~\cite{Xie_2017_resnext} 
and ShuffleNetV2~\cite{Ma_2018_ECCV_shufflenetv2}.
In this attack, we limit the query budget to $50k$ queries.
Similar to~\cite{mazeika2022steer,khaled2023} in training, we use an SGD optimizer with an initial learning rate of $0.01$ that is annealed with a cosine schedule. We use a Nesterov Momentum of $0.9$ and a weight decay of $5\cdot10^{-4}$. 
In data-free model stealing attacks DFME~\cite{truong2021data} and MAZE~\cite{kariyappa2021maze}, we choose ResNet-34-8x as the adversary model's architecture.
Similar to~\cite{truong2021data}, we train each model for $200$ epochs with $20M$ queries (i.e., generated images) using SGD with an initial learning rate of $0.1$ and a weight-decay of $5\cdot10^{-4}$. 
Furthermore, a scheduler was used to increase the learning rate by $0.3$ for $0.1\times$, $0.3\times$, and $0.5\times$ of the overall training period.
% \kacem{
We use Pytorch for all experiments including the quantization. We run our code on AlmaLinux 9 computer with 1 GPU NVIDIA Tesla P100-PCIE-12GB and 24 CPU cores Intel E5-2650 v4 Broadwell @~2.2GHz.
% }

% To have a fair evaluation, we run all experiments on a Linux CentOS~7 computer with 1 GPU NVIDIA Tesla P100-PCIE-12GB and 24 CPU cores Intel E5-2650 v4 Broadwell @~2.2GHz.

% GPU:  Tesla P100-PCIE-12GB
% CPU 24	 Intel E5-2650 v4 Broadwell @ 2.2GHz		NVIDIA P100 Pascal (12G HBM2 memory)

%% file: src/results.tex
\begin{table*}[!t]
\centering
\caption{Experimental results of multiple stealing attacks against quantized models. \\The variation between the original Large model and quantized models is reported between parentheses.\\
For RS, DCP and DivQAT, values are selected with $\ell_1$ distance lower than $0.6$.
}
\label{tab:results}
\begin{tabular}{@{}cccccc@{}}
\toprule
\multirow{2}{*}{Dataset} &
  % \multirow{2}{*}{\begin{tabular}[c]{@{}c@{}} Quantization \\algorithm \end{tabular}} &
  \multirow{2}{*}{\begin{tabular}[c]{@{}c@{}} Victim \\ Model \end{tabular}} &
  \multirow{2}{*}{\begin{tabular}[c]{@{}c@{}}Defender's
  classification\\ error  (victim model)\end{tabular}} &
  \multicolumn{3}{c}{Adversary's classification error (stolen model)} \\ \cmidrule(l){4-6} 
 &   &   &
  Knockoffnets Attack & DFME Attack &  MAZE Attack \\ 
  \midrule
						
  \multirow{5}{*}{CIFAR10}	& Large	 & 16.37	 & 20.56	 & 42.82	 & 67.70	\\
\cmidrule(l){2-6} 						
	& PTQ	 & 16.69 (+0.32)	 & 21.17 (+0.61)	 & 43.18 (+0.36)	 & 74.12 (+6.42)	\\
\cmidrule(l){2-6}

	& QAT	 & 15.52 (-0.85)	 & 20.40 (-0.16)	 & 48.74 (+5.92)	 & 82.97 (+15.27)	\\
\cmidrule(l){2-6}

	& RS (QAT)	 & 15.52 (-0.85)	 & 21.77 (+1.20)	 & 46.80 (+3.98)	 & 77.61 (+9.91)	\\
\cmidrule(l){2-6}

	& DCP (QAT)	 & 15.55 (-0.82)	 & 24.21 (+3.65)	 & 67.98 (+25.16)	 & 72.00 (+4.30)	\\
\cmidrule(l){2-6}

	& DivQAT	 & 20.03 (+3.66)	 & 23.52 (+2.96)	 & 69.40 (+26.58)	 & 90.00 (+22.30)	\\
\midrule						
\multirow{5}{*}{CIFAR100}	& Large	 & 39.68	 & 63.98	 & 85.93	 & 93.56	\\
\cmidrule(l){2-6} 						
	& PTQ	 & 40.25 (+0.57)	 & 63.94 (-0.04)	 & 85.06 (-0.87)	 & 95.17 (+1.61)	\\
\cmidrule(l){2-6}

	& QAT	 & 36.33 (-3.35)	 & 58.52 (-5.46)	 & 86.02 (+0.09)	 & 97.84 (+4.28)	\\
\cmidrule(l){2-6}

	& RS (QAT)	 & 36.33 (-3.35)	 & 57.76 (-6.22)	 & 86.88 (+0.95)	 & 94.72 (+1.16)	\\
\cmidrule(l){2-6}

	& DCP (QAT)	 & 36.35 (-3.33)	 & 65.28 (+1.29)	 & 87.03 (+1.10)	 & 95.87 (+2.31)	\\
\cmidrule(l){2-6}

	& DivQAT	 & 42.38 (+2.70)	 & 65.29 (+1.30)	 & 95.69 (+9.76)	 & 98.20 (+4.64)	\\
\midrule						
\multirow{5}{*}{GTSRB}	& Large	 & 7.73	 & 43.90	 & 35.97	 & 73.30	\\
\cmidrule(l){2-6} 						
	& PTQ	 & 7.76 (+0.03)	 & 50.18 (+6.29)	 & 34.87 (-1.10)	 & 83.95 (+10.65)	\\
\cmidrule(l){2-6}

	& QAT	 & 7.23 (-0.50)	 & 50.18 (+6.29)	 & 34.87 (-1.10)	 & 83.95 (+10.65)	\\
\cmidrule(l){2-6}

	& RS (QAT)	 & 7.23 (-0.50)	 & 46.31 (+2.41)	 & 36.52 (+0.55)	 & 78.80 (+5.49)	\\
\cmidrule(l){2-6}

	& DCP (QAT)	 & 7.23 (-0.50)	 & 51.71 (+7.81)	 & 44.58 (+8.61)	 & 82.02 (+8.72)	\\
\cmidrule(l){2-6}

	& DivQAT	 & 6.88 (-0.85)	 & 52.03 (+8.14)	 & 64.76 (+28.79)	 & 93.25 (+19.95)	\\
\midrule							
\multirow{5}{*}{SVHN}	& Large	 & 5.82	 & 9.22	 & 6.89	 & 7.18	\\
\cmidrule(l){2-6} 						
	& PTQ	 & 5.87 (+0.05)	 & 9.22 (+0.00)	 & 6.48 (-0.41)	 & 49.90 (+42.72)	\\
\cmidrule(l){2-6}

	& QAT	 & 5.78 (-0.04)	 & 9.22 (+0.00)	 & 6.64 (-0.25)	 & 6.56 (-0.63)	\\
\cmidrule(l){2-6}

	& RS (QAT)	 & 5.78 (-0.04)	 & 9.52 (+0.30)	 & 5.93 (-0.96)	 & 32.45 (+25.27)	\\
\cmidrule(l){2-6}

	& DCP (QAT)	 & 5.78 (-0.04)	 & 18.43 (+9.21)	 & 13.50 (+6.61)	 & 18.04 (+10.86)	\\
\cmidrule(l){2-6}

	& DivQAT	 & 9.28 (+3.46)	 & 20.97 (+11.75)	 & 84.06 (+77.17)	 & 92.24 (+85.06)	\\
\bottomrule										

\end{tabular}
\end{table*}

\begin{table}[!t]
\centering
\caption{Impact of combining DivQAT with post-training defenses. \\The variation between the QAT and DivQAT is reported between parentheses.\\
For RS, DCP and DivQAT, values are selected with $\ell_1$ distance lower than $0.6$.
}
\label{tab:results-combination}
\begin{tabular}{@{}cccc@{}}
\toprule
\multirow{3}{*}{Dataset} &
  % \multirow{2}{*}{\begin{tabular}[c]{@{}c@{}} Quantization \\algorithm \end{tabular}} &
  \multirow{3}{*}{\begin{tabular}[c]{@{}c@{}} Victim \\ Model \end{tabular}} &
  \multirow{3}{*}{\begin{tabular}[c]{@{}c@{}} Defender's \\
  classification error  \\ (victim model)\end{tabular}} &
  \multirow{3}{*}{\begin{tabular}[c]{@{}c@{}} Adversary's \\ classification error \\ (stolen model)\end{tabular}}
  \\
  &   &   &  \\
  &   &   &  \\
  \midrule
									
\multirow{4}{*}{CIFAR10}	& QAT+RS	 & 15.52	 & 21.77	\\

\cmidrule(l){2-4}

	& DivQAT+RS	 & 16.41 (+0.89)	 & 24.27 (+2.51)	\\
\cmidrule(l){2-4}

	& QAT+DCP	 & 15.55	 & 24.21	\\
\cmidrule(l){2-4}

	& DivQAT+DCP	 & 16.24 (+0.69)	 & 26.88 (+2.67)	\\
\midrule

		\multirow{4}{*}{CIFAR100}	& QAT+RS	 & 36.33	 & 57.76	\\
\cmidrule(l){2-4}

	& DivQAT+RS	 & 36.65 (+0.32)	 & 61.25 (+3.48)	\\
\cmidrule(l){2-4}

	& QAT+DCP	 & 36.35	 & 65.28	\\
\cmidrule(l){2-4}

	& DivQAT+DCP	 & 36.87 (+0.52)	 & 69.46 (+4.18)	\\
\midrule				
\multirow{4}{*}{GTSRB}	& QAT+RS	 & 7.23	 & 46.31	\\
\cmidrule(l){2-4}

	& DivQAT+RS	 & 6.88 (-0.35)	 & 54.30 (+7.99)	\\
\cmidrule(l){2-4}

	& QAT+DCP	 & 7.23	 & 51.71	\\
\cmidrule(l){2-4}

	& DivQAT+DCP	 & 6.96 (-0.27)	 & 59.36 (+7.65)	\\
				
\midrule				
\multirow{4}{*}{SVHN}	& QAT+RS	 & 5.78	 & 9.52	\\
\cmidrule(l){2-4}

	& DivQAT+RS	 & 6.62 (+0.84)	 & 13.15 (+3.63)	\\
\cmidrule(l){2-4}

	& QAT+DCP	 & 5.78	 & 18.43	\\
\cmidrule(l){2-4}

	& DivQAT+DCP	 & 5.57 (-0.21)	 & 20.56 (+2.13)	\\
\bottomrule				
\end{tabular}
\end{table}

% \begin{table*}[!t]
% \centering
% \caption{Experimental results of multiple stealing attacks against quantized models. \\The variation between QAT and DivQAT is reported between parentheses.}
% \label{tab:results}
% \begin{tabular}{@{}cccccc@{}}
% \toprule
% \multirow{2}{*}{Dataset} &
%   \multirow{2}{*}{\begin{tabular}[c]{@{}c@{}} Quantization \\algorithm \end{tabular}} &
%   \multirow{2}{*}{\begin{tabular}[c]{@{}c@{}}Defender's
%   classification\\ error  (victim model)\end{tabular}} &
%   \multicolumn{3}{c}{Adversary's classification error (stolen model)} \\ \cmidrule(l){4-6} 
%  &   &   &
%   Knockoffnets Attack & DFME Attack &  MAZE Attack \\ \midrule
% \multirow{2}{*}{CIFAR10} &
%   QAT (normal) &  16.41 &  20.99 &  57.30 &  80.99 \\ \cmidrule(l){2-6} 
%  &
%   DivQAT (ours) &
%   16.32 (-0.09) &
%   22.62 (+1.63) &
%   59.92 (+2.62) &
%   90.00 (+9.01) \\ \midrule
% \multirow{2}{*}{CIFAR100} &
%   QAT (normal) &  52.51 &  59.81 &  91.20 &  86.63 \\ \cmidrule(l){2-6} 
%  &
%   DivQAT (ours) &
%   52.43 (-0.08) &
%   60.99 (+1.17) &
%   92.63 (+1.43) &
%   99.00 (+12.37) \\ \midrule
% \multirow{2}{*}{GTSRB} &
%   QAT (normal) &  7.05 &  49.05 &  34.90 &  82.07 \\ \cmidrule(l){2-6} 
%  &
%   DivQAT (ours) &
%   7.07 (+0.02) &
%   54.33 (+5.28)  &% {52.46 (+3.41)}
%   54.54 (+19.64) &
%   99.52 (+17.45) \\ \midrule
% \multirow{2}{*}{SVHN} &
%   QAT (normal) &  5.26 &  9.22 &  5.68 &  5.65 \\ \cmidrule(l){2-6} 
%  &
%   DivQAT (ours) &
%   5.24 (-0.02) &
%   10.01 (+0.79) &
%   5.81 (+0.13) &
%   93.30 (+87.65) \\ \bottomrule
% \end{tabular}
% \end{table*}

\begin{figure}[!t]
\centering
\subfloat[\label{fig:archs-clf-err}]{%
\includegraphics[width=0.5\textwidth]{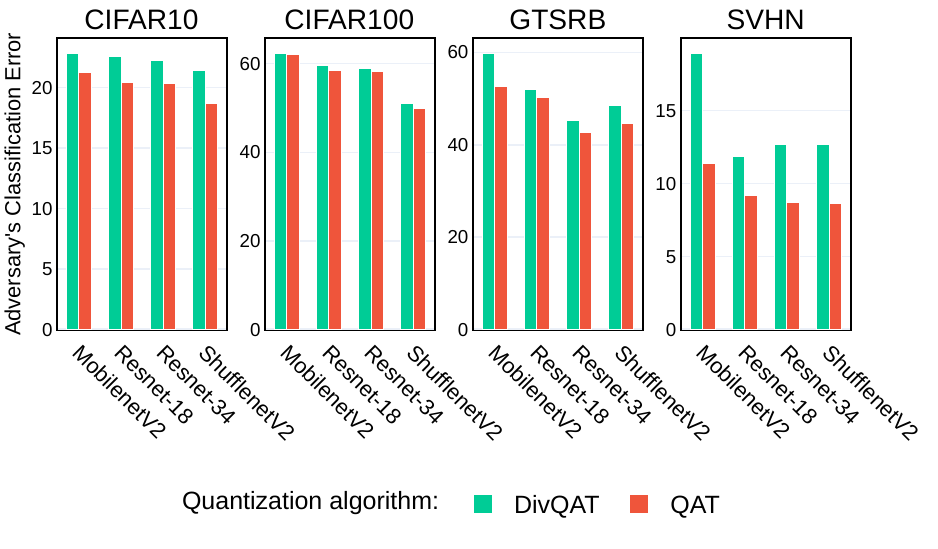}
} \hfil
\subfloat[\label{fig:archs-disagreement}]{%
\includegraphics[width=0.5\textwidth]{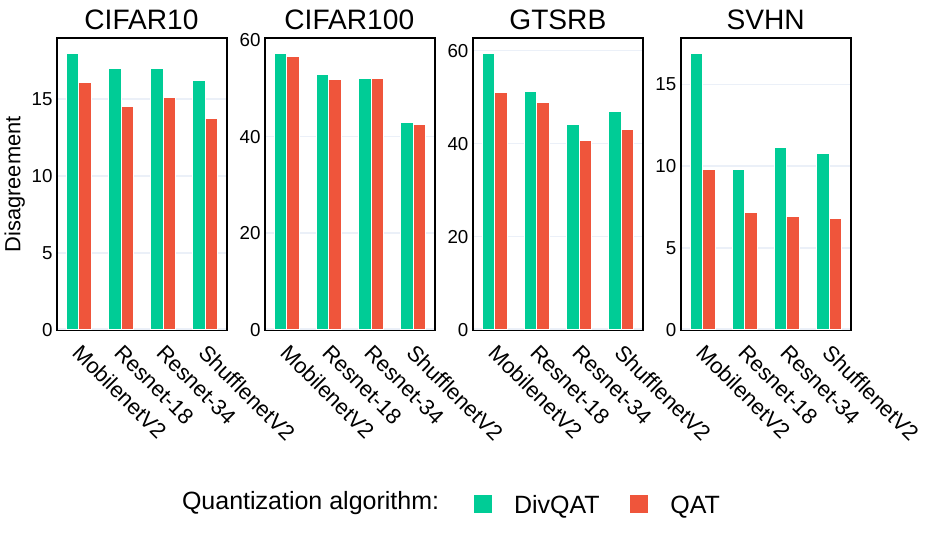}
} \hfil
\caption{Results of Knockoffnets attack using different adversary architectures to extract quantized models obtained with QAT and DivQAT. (a) and (b) respectively show the adversary's classification error and the disagreement between the victim and the adversary. Each column represents the results for a dataset. Architectures are on the x-axis.}
\label{fig:archs}
\end{figure}

% \begin{figure*}[!t]
% \centering
% \subfloat[\label{fig:disagreement}]{%
% \includegraphics[width=1\textwidth]{figures/extraction--defender_err-disagreement.pdf}
% } \hfil
% \subfloat[\label{fig:clf-error}]{%
%   \includegraphics[width=\textwidth]{figures/extraction--defender_err-adversary_err.pdf}%
% } \hfil
% \subfloat[\label{fig:l1-distance}]{%
% \includegraphics[width=1\textwidth]{figures/extraction--l1_distance-adversary_err.pdf}
% } \hfil
% \caption{Imapct of combining our quantization algorithm DivQAT with RS and DCP defenses against Knockoffnets stealing attack. 
% Curves are obtained by varying the defenses (RS and DCP) parameters. 
% The y-axis shows the Disagreement in (a) and the Adversary's Classification Error in (b) and (c). 
% The Defender's Classification Error on the x-axis in (a) and (b) represents the victim model's performance. 
% The $\ell_1$ distance in (c) represents the noise magnitude injected in the model's prediction probabilities by each defense technique. Each column represents results on a dataset. Combining a defense with DivQAT shows better performance than a combination with QAT.
% % ``$\uparrow$'' means higher values are better, and ``$\downarrow$'' means lower values are better.
% % \textcolor{cyan}{maybe put DCP:dashed and RS: pointed, for easier visual comparaison + maybe explain if one curve is above of the other, which one is better?}
% }
% \label{fig:adv-error}
% \end{figure*}

\subsection{Experimental results}

\subsubsection{DivQAT robustness against extraction attacks}
First, we evaluate the robustness of quantized models obtained by our algorithm DivQAT compared to
% \kacem{
an extraction attack on the original \emph{large} model and on
other models obtained through quantization techniques QAT and PTQ as well as quantized models protected with post-training defenses RS~\cite{lee2018defending} and DCP~\cite{khaled2023}.
Table~\ref{tab:results} shows the results of 3 stealing attacks KnockoffNets~\cite{orekondy2019knockoff}, DFME~\cite{truong2021data} and MAZE~\cite{kariyappa2021maze}. 
% }
For each attack, we report the performance of the victim and the corresponding stolen model using the classification error as an evaluation metric. We aim to maximize this metric for the adversary's model (i.e., the stolen model), and keep its increase as small as possible for the defender's model (i.e., the victim).
% \kacem{
Since defenses against extraction attacks add perturbation to the prediction probabilities, this may harm the confidence values of the model.
Therefore, we constrain the results of RS, DCP and DivQAT to have a budget of $\ell_1$ distance between their predictions and ones of a model obtained through normal QAT to be lower than 
$0.6$, which is a reasonable value to keep the noise as small as possible. 
This ensures the transparency of models about the precision of their probabilities.
% }

Compared to normal QAT, our algorithm DivQAT, yields a defender's neural network with a comparable performance. In most cases, the classification error variation of victim models is negligible with an utmost performance drop of~$3.66\%$.
Furthermore, DivQAT deteriorates the stolen model performance in all attack settings. The classification error of the adversary models is increased by $1.30\%$ to $11.75\%$ for KnockoffNets attacks, by $9.76\%$ to $77.17\%$ for DFME attacks, and by $4.64\%$ to $85.06\%$ for MAZE attacks. 
We observe that DFME and MAZE attacks are the most harmed by our defense since these attacks rely on the victim's predictions to generate suitable images that will be used to query the victim model and improve its extraction. This entails that our approach poisons the data generated by these models and hinders the adversary's learning process.

We notice that KnockoffNets is the strongest attack because in most cases it produces a stolen model with the lowest classification error. Additionally, the error increase induced by our DivQAT algorithm compared to QAT is small for KnockoffNets compared to the other attacks. We experiment with this attack further using different CNN architectures
ResNet-18, ResNet-34, MobileNetV2~\cite{Sandler_2018_MobileNetV2}, 
% ResNeXt-29~\cite{Xie_2017_resnext} 
and ShuffleNetV2~\cite{Ma_2018_ECCV_shufflenetv2}.
Fig.~\ref{fig:archs} shows the results of Knockoffnets attacks against quantized models. For each attack, a different adversary architecture is selected. Our algorithm shows an increase in the adversary's classification error and the disagreement for most cases. 
We notice that the lightweight network MobileNetV2 often yields a higher error compared to other architectures, which proves that our quantization technique makes it harder for weak adversaries to learn from perturbed outputs compared to larger models.
% This suggests add
% \kacem{so, all in all, this further investigation into Knockoffnets resulted in what ? }
% \kacem{next}

\subsubsection{Impact of combining DivQAT with other defenses against extraction attacks}

We explore the scenario where a model owner using our quantization technique wants to add an extra layer of protection against extraction attacks using state-of-the-art defenses suitable for quantized models.
Table~\ref{tab:results-combination} shows the results of such a scenario where quantized models are attacked with Knockoffnets attack, the strongest attack in previous experiments.
Attacks are performed against multiple variations of defended models. Each defense technique has a set of parameters that can be adjusted to balance the performance trade-off between the defender and the adversary:
an increase in the adversary's classification error comes with the cost of an increase in the defender's classification error.
Like previous experiments, to balance both metrics we constrain the models with a perturbation level measured by $\ell_1$ distance that should not exceed $0.6$.
Models quantized with DivQAT show a significantly improved robustness against extraction attacks compared to models quantized with normal QAT. 
% In Fig.~\ref{fig:disagreement} and Fig.~\ref{fig:clf-error},  
We observe that 
% both 
the \textit{Adversary's classification error} 
% and the \textit{Disagreement} have
has increased in all cases. 
For example,  for the GTSRB dataset, the classification error
% and the disagreement are 
is up by $\sim8\%$ for both  DivQAT$+$RS and DivQAT$+$DCP  compared to respectively QAT$+$RS and QAT$+$DCP. 

% is
% up by $7.65\%$ for the DivQAT$+$DCP compared to QAT$+$DCP. For the SVHN dataset, the classification error 
% and the disagreement are 

% \kacem{
We investigate the impact of each defense on the prediction probabilities. RS and DCP are perturbation-based defenses, hence their efficacity relies on noising the predictions. 
The defender aims to maximize the adversary's error while minimally perturbing the predictions. 
Table~\ref{tab:results} and Table~\ref{tab:results-combination} show the adversary's classification error with respect to a fixed budget of $\ell_1$ distance between clean and perturbed probabilities. 
We observe that models trained with DivQAT algorithm show better performance compared to other quantized models and baseline defenses. 
% For example, for a $\ell_1$ distance of $0.6$, the adversary's classification error can be increased by up to $\sim9\%$ for the GTSRB dataset in the DivQAT$+$RS experiment. 
This entails that when using DivQAT combined with other perturbation-based defenses, the robustness against stealing attacks is enhanced while the injected noise in the prediction probabilities is reduced.
% }

\subsubsection{Performance of DivQAT with different quantization configurations}
% \kacem{
Table~\ref{tab:results-server-mobile} 
shows the results of KnockoffNets attack against quantized models obtained through multiple quantization techniques. For each approach QAT and DivQAT, we use two quantization variations \emph{server} and \emph{mobile}. 
The perturbation level is constrained not to exceed the $\ell_1$ distance of $0.6$.
We notice that in both scenarios, models quantized using our technique DivQAT yield better robustness against extraction attacks compared to QAT.
This validates the efficacy of our approach in protecting quantized models while having a minor accuracy drop and minimal perturbation levels.
% }

\begin{table}[!t]
\centering
\caption{Extraction attacks against different quantization configurations \\The variation between QAT and DivQAT is reported between parentheses.\\
For DivQAT, values are selected with $\ell_1$ distance lower than $0.6$.
}
\label{tab:results-server-mobile}
\begin{tabular}{@{}cccc@{}}
\toprule
\multirow{3}{*}{Dataset} &
  % \multirow{2}{*}{\begin{tabular}[c]{@{}c@{}} Quantization \\algorithm \end{tabular}} &
  \multirow{3}{*}{\begin{tabular}[c]{@{}c@{}} Victim \\ Model \end{tabular}} &
  \multirow{3}{*}{\begin{tabular}[c]{@{}c@{}} Defender's \\
  classification error  \\ (victim model)\end{tabular}} &
  \multirow{3}{*}{\begin{tabular}[c]{@{}c@{}} Adversary's \\ classification error \\ (stolen model)\end{tabular}}
  \\
  &   &   &  \\
  &   &   &  \\
  \midrule
  \multirow{4}{*}{CIFAR10}	& QAT (server)	 & 15.52	 & 20.40	\\
				
\cmidrule(l){2-4} 				
	& DivQAT (server)	 & 20.03 (+4.51)	 & 23.52 (+3.12)	\\
\cmidrule(l){2-4} 				
	& QAT (mobile)	 & 15.51	 & 21.37	\\
\cmidrule(l){2-4} 				
	& DivQAT (mobile)	 & 18.86 (+3.35)	 & 24.12 (+2.76)	\\
\midrule				
\multirow{4}{*}{CIFAR100}	& QAT (server)	 & 36.33	 & 58.52	\\
				
\cmidrule(l){2-4} 				
	& DivQAT (server)	 & 42.38 (+6.05)	 & 65.29 (+6.77)	\\
\cmidrule(l){2-4} 				
	& QAT (mobile)	 & 36.37	 & 58.97	\\
\cmidrule(l){2-4} 				
	& DivQAT (mobile)	 & 37.60 (+1.23)	 & 61.88 (+2.92)	\\
\midrule				

\multirow{4}{*}{GTSRB}	& QAT (server)	 & 7.23	 & 50.18	\\
				
\cmidrule(l){2-4} 				
	& DivQAT (server)	 & 6.88 (-0.35)	 & 52.03 (+1.85)	\\
\cmidrule(l){2-4} 				
	& QAT (mobile)	 & 7.28	 & 51.65	\\
\cmidrule(l){2-4} 				
	& DivQAT (mobile)	 & 6.62 (-0.67)	 & 53.55 (+1.89)	\\
\midrule				
\multirow{4}{*}{SVHN}	& QAT (server)	 & 5.78	 & 9.22	\\
				
\cmidrule(l){2-4} 				
	& DivQAT (server)	 & 9.28 (+3.50)	 & 20.97 (+11.75)	\\
\cmidrule(l){2-4} 				
	& QAT (mobile)	 & 5.74	 & 9.22	\\
\cmidrule(l){2-4} 				
	& DivQAT (mobile)	 & 9.37 (+3.63)	 & 14.92 (+5.70)	\\
\bottomrule				
\end{tabular}
\end{table}

% \begin{figure*}[]
% \centering
% \includegraphics[width=1\textwidth]{figures/extraction--l1_distance-adversary_err.pdf}
% \caption{l1 distance}
% \label{fig:l1-distance}
% \end{figure*}

% \begin{figure*}[]
% \centering
% \includegraphics[width=1\textwidth]{figures/extraction--defender_err-disagreement.pdf}
% \caption{Disagreement}
% \label{fig:disagreement}
% \end{figure*}

\subsubsection{Ablation study}
\begin{figure}[!t]
\centering
\includegraphics[width=0.5\textwidth]{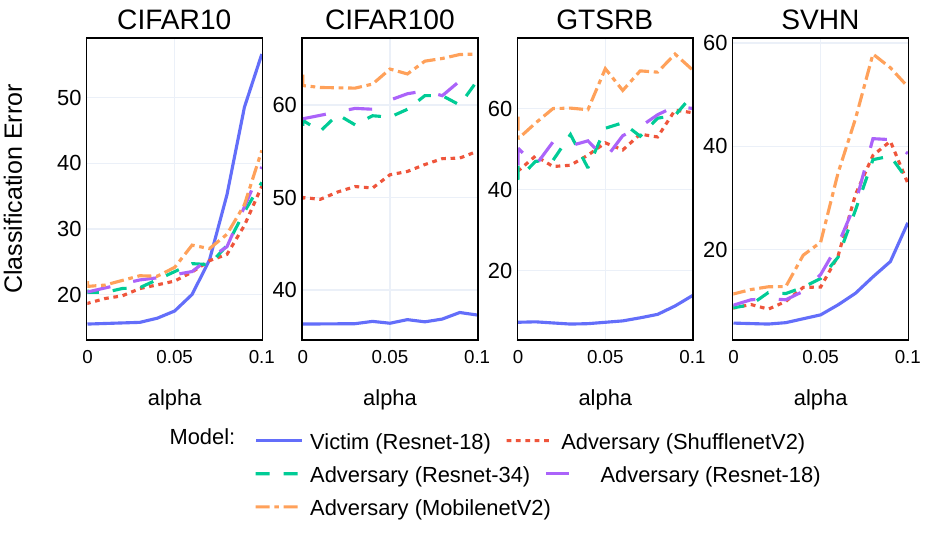}
\caption{
Impact of $\alpha$ variation in DivQAT on the classification error of both the victim and the stolen model (adversary).
% Architectures are on the x-axis.
}
\label{fig:alpha-error}
\end{figure}

\begin{figure}[!t]
\centering
\includegraphics[width=0.5\textwidth]{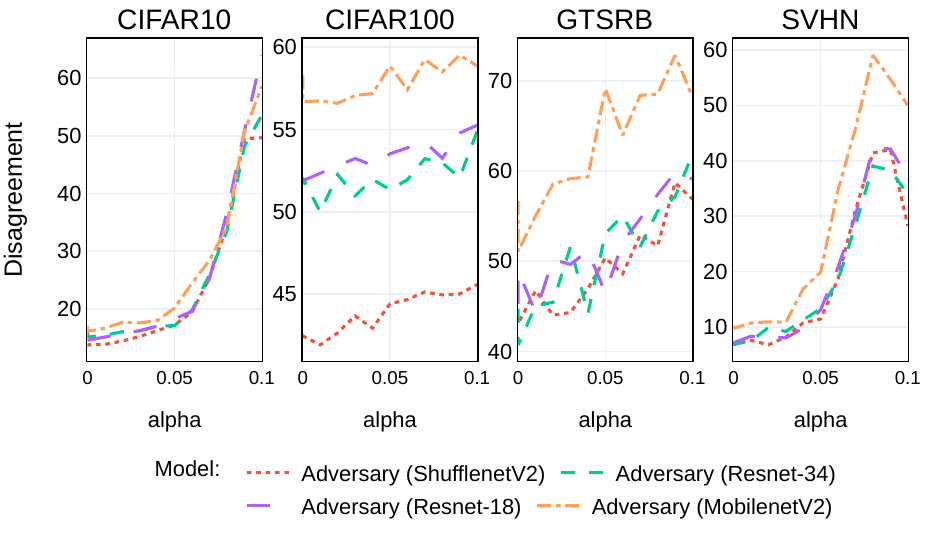}
\caption{
Impact of $\alpha$ variation in DivQAT on the disagreement between the victim and the adversary. 
% Architectures are on the x-axis.
}
\label{fig:alpha-disagreement}
\end{figure}

% \kacem{
We investigate the impact of the parameter $\alpha$  in Eq.\ref{eq:loss_div_qat} on the robustness of the quantized model against extraction attacks.
Fig.~\ref{fig:alpha-error} and Fig.~\ref{fig:alpha-disagreement}  show the results of this study.
We quantize multiple models using DivQAT with different $\alpha$ values, then each model is extracted using multiple CNN architectures as adversary models, namely ResNet-18, ResNet-34, MobileNetV2~\cite{Sandler_2018_MobileNetV2}, 
% ResNeXt-29~\cite{Xie_2017_resnext} 
and ShuffleNetV2~\cite{Ma_2018_ECCV_shufflenetv2}.
Our experiments show that increasing $\alpha$ results in a higher classification error for both the victim and the adversary. Additionally, for high $\alpha$ values,  the disagreement between the victim and the adversary is larger.
This entails that $\alpha$ is a trade-off parameter that balances robustness against extraction attacks and the defender's performance. 
% }

\begin{figure}[!t]
\centering
\includegraphics[width=0.5\textwidth]{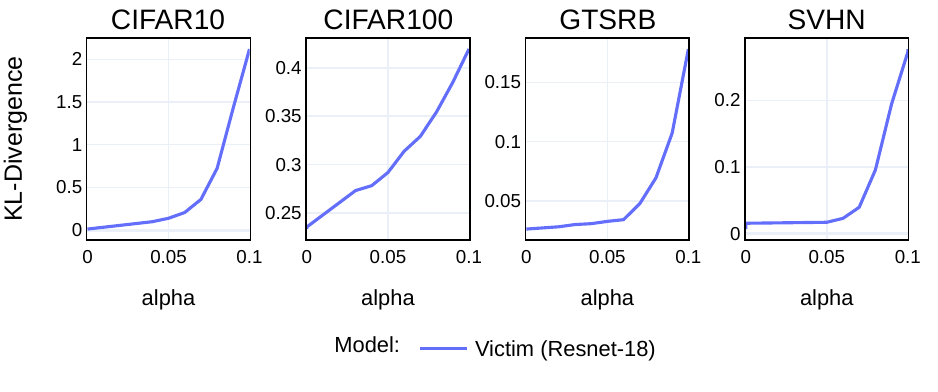}
\caption{
Impact of $\alpha$ variation in DivQAT on the KL-Divergence between the large and quantized model.
}
\label{fig:alpha-kl-div}
\end{figure}

% \kacem{
Since our quantization approach relies on maximizing the KL-divergence between the original \emph{large} model and its \emph{quantized} counterpart, we study the variation of the KL-Divergence between both models w.r.t $\alpha$.
Fig.~\ref{fig:alpha-kl-div} shows the result of this experiment.
We find that increasing $\alpha$ results in increasing the KL-Divergence between the large and quantized models.
This empirically proves our intuition in Eq.~\ref{eq:loss_div_qat} that through our quantization algorithm, we obtain a quantized model slightly divergent from the original one. 

\subsection{Discussion}
Our experimental results underscore the robustness of the DivQAT method against model extraction attacks, demonstrating its superiority over traditional QAT and other defenses applicable to quantized models. 
Notably, this enhanced security does not compromise the performance of the victim model, which only has a minor decrease in accuracy. 
When compared to existing quantization methods and baseline defenses, DivQAT stands out for its ability to significantly increase the adversary's classification error and the disagreement between the defender and the adversary, thereby offering a more secure model against extraction attacks. 
% \kacem{
These findings open up new avenues for future research, particularly in terms of focusing on the practicality and efficiency of defense mechanisms against extraction attacks, as well as 
integrating 
these defenses during the quantization process. 
This also suggests revisiting quantization algorithms to create models that are robust by design.
% }
% DivQAT with other defense mechanisms and applying it across a variety of neural network architectures and datasets. 

However, it's important to acknowledge the limitations of DivQAT as it is effective against attackers that require access to prediction probabilities. In other scenarios, these prediction probabilities might not be available and the victim model only provides the top-1 class label~\cite{Zhou_2020_CVPR_dast,Sanyal_2022_CVPR_hard}.
Therefore, deviating the probability distribution might not be an effective strategy against attacks that steal the model without relying on the probability values. 
Additionally, the scope of our paper only involves the quantization of \ac{CNNs} for computer vision tasks, thus the impact of DivQAT on tasks other than image classifications (e.g., natural language
processing) and its effectiveness is an area that can be explored in future research directions. 
% As such, our future research will focus on these aspects to fully realize the potential of DivQAT in securing models against extraction attacks.w

%% file: src/conclusion.tex
In conclusion, this paper presents DivQAT, a robust \ac{QAT} technique designed to protect \ac{DL} models on edge devices from model extraction attacks. Our extensive experiments on benchmark vision datasets have demonstrated the effectiveness of DivQAT in enhancing model robustness without compromising accuracy. Furthermore, DivQAT has shown potential in not only defending models against direct attacks but also in enhancing the effectiveness of other defense mechanisms, thereby providing a comprehensive, multi-layered security approach. 
Our approach ensures that security considerations are integrated into the model development process, rather than being an afterthought.
This paper paves the way for future research into optimizing quantization techniques for improved security and efficiency in edge computing applications.